\documentclass[sigconf, nonacm]{acmart}

\usepackage{booktabs,tabularx,makecell,array}
\newcolumntype{L}[1]{>{\raggedright\arraybackslash}p{#1}} 
\newcolumntype{Y}{>{\raggedright\arraybackslash}X} 

\AtBeginDocument{%
  }

\begin{document}

\title{SafeHumanoid: VLM-RAG-driven Control of Upper Body Impedance for Humanoid Robot}


\author{Yara Mahmoud}
\affiliation{%
  \institution{Skolkovo Institute of Science and Technology}
  \city{Moscow}
  \country{Russia}}
\email{Yara.Mahmoud@skoltech.ru}

\author{Jeffrin Sam}
\affiliation{%
  \institution{Skolkovo Institute of Science and Technology}
  \city{Moscow}
  \country{Russia}}
\email{Jeffrin.Sam@skoltech.ru}

\author{Nguyen Khang}
\affiliation{%
  \institution{Skolkovo Institute of Science and Technology}
  \city{Moscow}
  \country{Russia}}
\email{Nguyen.Khang@skoltech.ru}

\author{Marcelino Fernando}
\affiliation{%
  \institution{Skolkovo Institute of Science and Technology}
  \city{Moscow}
  \country{Russia}}
\email{Marcelino.Fernando@skoltech.ru}

\author{Issatay Tokmurziyev}
\affiliation{%
  \institution{Skolkovo Institute of Science and Technology}
  \city{Moscow}
  \country{Russia}}
\email{Issatay.Tokmurziyev@skoltech.ru}

\author{Miguel Altamirano Cabrera}
\affiliation{%
  \institution{Skolkovo Institute of Science and Technology}
  \city{Moscow}
  \country{Russia}}
\email{m.altamirano@skoltech.ru}

\author{Muhammad Haris Khan}
\affiliation{%
  \institution{Skolkovo Institute of Science and Technology}
  \city{Moscow}
  \country{Russia}}
\email{Haris.Khan@skoltech.ru}

\author{Artem Lykov}
\affiliation{%
  \institution{Skolkovo Institute of Science and Technology}
  \city{Moscow}
  \country{Russia}}
\email{Artem.Lykov@skoltech.ru}

\author{Dzmitry Tsetserukou}
\affiliation{%
  \institution{Skolkovo Institute of Science and Technology}
  \city{Moscow}
  \country{Russia}}
\email{d.tsetserukou@skoltech.ru}


\renewcommand{\shortauthors}{Y. Mahmoud et al.}

\begin{abstract}
Safe and trustworthy Human–Robot Interaction (HRI) requires robots not only to complete tasks but also to regulate impedance and speed according to scene context and human proximity. We present SafeHumanoid, an egocentric vision pipeline that links Vision–Language Models (VLMs) with Retrieval-Augmented Generation (RAG) to schedule impedance and velocity parameters for a humanoid robot. Egocentric frames are processed by a structured VLM prompt, embedded and matched against a curated database of validated scenarios, and mapped to joint-level impedance commands via inverse kinematics. We evaluate the system on tabletop manipulation tasks with and without human presence, including wiping, object handovers, and liquid pouring. The results show that the pipeline adapts stiffness, damping, and speed profiles in a context-aware manner, maintaining task success while improving safety. Although current inference latency (up to 1.4 s) limits responsiveness in highly dynamic settings, SafeHumanoid demonstrates that semantic grounding of impedance control is a viable path toward safer, standard-compliant humanoid collaboration.
\end{abstract}







\maketitle

\begin{figure}[t!]
\centerline{\includegraphics[width=0.5\textwidth]{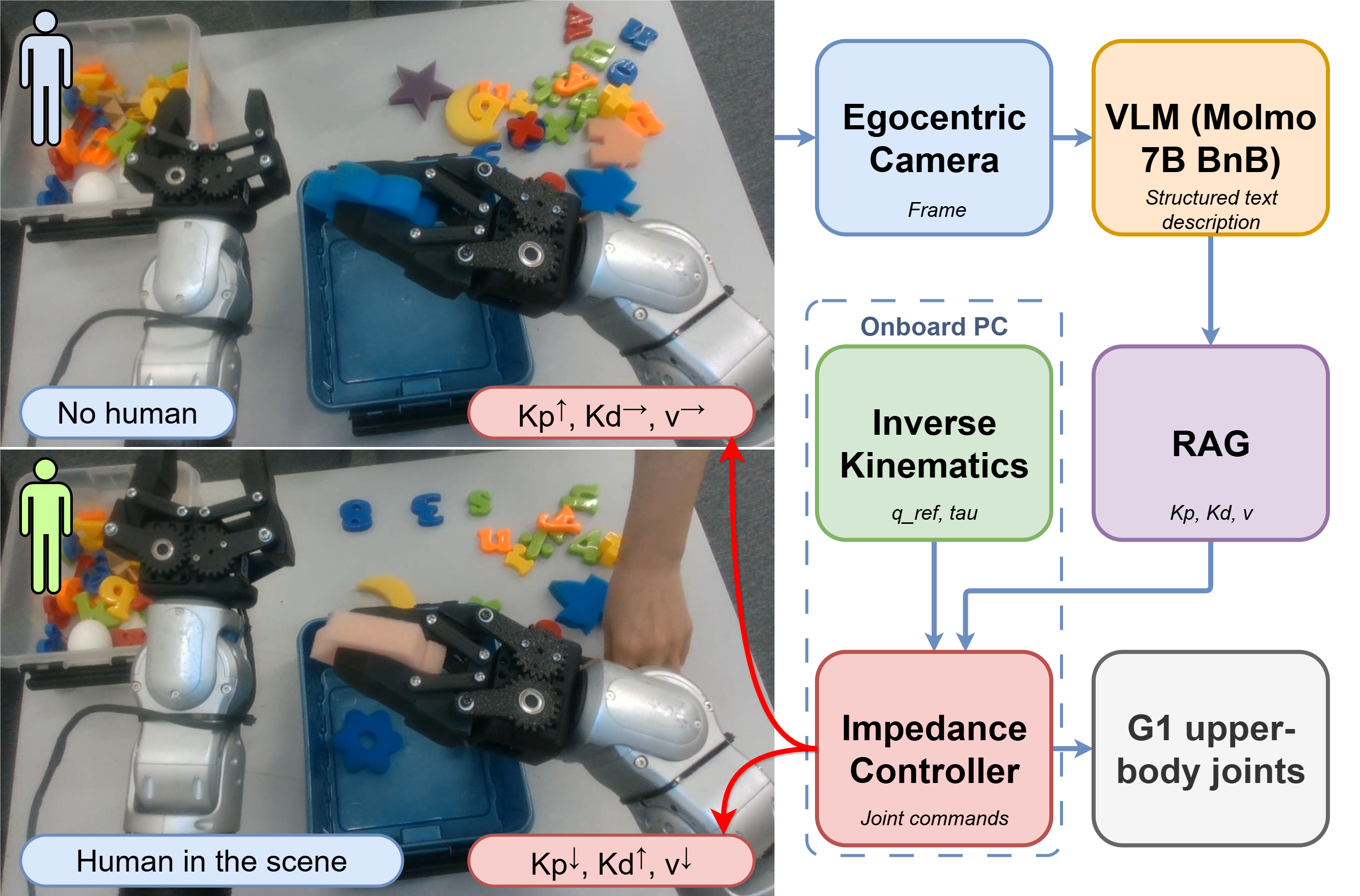}}
\caption{Egocentric perception and semantic-to-safety pipeline. Left: robot grasping with and without human presence, showing adaptive modulation of $K_p$, $K_d$, and $v$. Right: high-level flow from camera input through VLM–RAG reasoning to impedance control of G1 upper-body joints.}
\label{fig:first}
\vspace{-0.5cm}
\end{figure}

\section{Introduction}

Robots are moving from industrial cells into shared human spaces--homes, workshops, and hospitals--where success is no longer defined by task completion alone but also by guarantees on safety and comfort. Safe Human–Robot Interaction (HRI) requires predictable, gentle contact with bounded speed and force~\cite{ref1}.

A key part of the solution for years has been impedance control, which allows robots to contact with a programmed sense of stiffness and damping~\cite{ref2}. This was a major leap beyond rigid, position-controlled machines. Yet, a robot with a single, fixed impedance setting struggles to handle the sheer variety of human--robot interaction. The soft touch needed to hand someone a fragile object is fundamentally different from the firm stability required to wipe down a table. Adaptive controllers that learn from contact or adjust reactively have been explored~\cite{ref3,ref4}, but they often respond only after interaction, rather than proactively ensuring safe behaviour.

Vision enables anticipation. Using vision, robots can plan their paths to be less intrusive~\cite{ref5} and can slow down when people draw near, a principle echoed in modern collaborative robot safety standards~\cite{ref6}. This is a crucial capability, but it reveals a deeper limitation: these systems are often geometrically aware but semantically blind. They can register a person’s location, but have no understanding of context. A robot that sees a person standing two feet away has no way of knowing if that person is a passive observer or an active collaborator. Without contextual understanding, their responses remain guesswork.

Recent advances in Vision--Language--Models (VLMs) and Vision--Language--Action (VLA) models have opened a new door. These models give robots something they have historically lacked: a measure of common sense and reasoning~\cite{ref7}. Already, VLAs are being used to translate complex, high-level commands into task sequences~\cite{ref8,ref9} and joint trajectories~\cite{refX}. However, even when trained end-to-end to generate robot motions, such models typically focus on positional control and task completion. They do not inherently account for impedance behaviour---the modulation of stiffness, damping, and speed that governs safe physical interaction. As a result, robots may execute the correct sequence of joint positions while still lacking the compliance and safety guarantees required in human--robot shared environments.

Current VLM/VLA pipelines focus on task feasibility and positional execution; they rarely consider how motions should be executed safely in close proximity to humans. This gap motivates the need for a framework that grounds semantic reasoning in impedance and compliance. Recent efforts such as ~\cite{khan2025shakevlavisionlanguageactionmodelbasedbimanual} and ~\cite{batool2025impedancegptvlmdrivenimpedancecontrol} show that combining VLMs with Retrieval-Augmented Generation (RAG) can improve grounding in manipulation and drone navigation, but such approaches have not yet been explored in humanoid HRI.

Our work, \textit{SafeHumanoid}, tackles this problem head-on by creating a bridge between the semantic reasoning of VLMs and the requirements of safe and efficient impedance control. Using egocentric vision, the VLM interprets task and scene context, while a RAG pipeline queries a curated database of standard-aligned parameters ($K_p$, $K_d$, $v$)~\cite{ref10}. This allows the robot to use its judgment to consult a curated database of safe, reliable control parameters, much like a human expert referencing a trusted manual. These are integrated with inverse kinematics that provide joint references and feed-forward torques for gravity compensation, yielding per-joint commands
\[
\{ q_{\text{ref}}, \ \dot{q}_{\text{ref}}, \ \tau_{\text{ff}}, \ K_p, \ K_d \}.
\]

Importantly, our approach is agnostic to how joint trajectories are generated. The desired positions could come from a VLA model, a task-planning framework such as Gr00t, or even a pose-estimation pipeline for object manipulation. What our method contributes is the missing compliance layer: dynamically scheduling impedance parameters so that any trajectory is executed in a way that aligns with human--robot safety standards and task requirements.

This high-level reasoning is then seamlessly integrated into a robust Upper-body control stack that accounts for the robot's dynamics, allowing it to move from simple reactions to nuanced, context-driven behaviours.

Prior systems are either purely geometric or focus on task-level
sequencing. What they lack is the integration of semantic reasoning into the safety and compliance layer — impedance and speed scheduling. Our novelty is precisely this bridge: semantics guiding impedance control

In this paper, we present the following contributions:
\begin{itemize}
    \item We introduce a VLM--RAG control pipeline that uses egocentric vision and task prompts to select safe and context-aware impedance and speed parameters in real time.
    \item We demonstrate the full system running on the Unitree G1 humanoid, showing a practical integration of high-level semantic reasoning with low-level upper-body motion control.
    \item We provide experimental results from manipulation and near-human tasks, showing that our method maintains task success while providing safer and more adaptive behaviour than fixed-gain baselines.
\end{itemize}
To provide an overview, Fig. ~\ref{fig:first} illustrates the SafeHumanoid pipeline, showing how egocentric vision is processed by the VLM–RAG reasoning stack and translated into impedance-aware joint control on the Unitree G1.

\begin{figure*}[t]
    \centering
    \includegraphics[width=\textwidth]{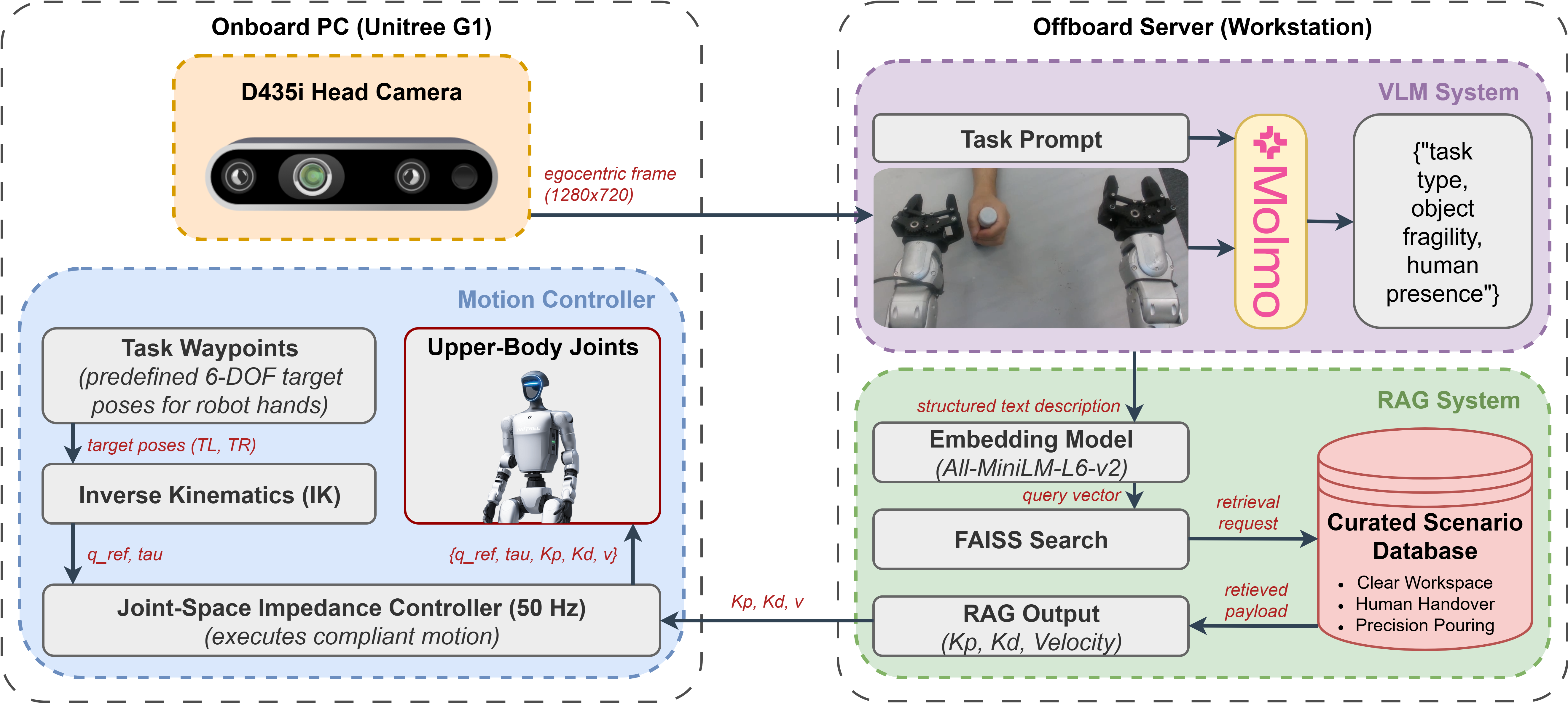} 
    \caption{SafeHumanoid pipeline architecture. The onboard PC streams egocentric frames and executes impedance control at 50 Hz, while the offboard workstation (Molmo VLM + FAISS-based RAG) grounds scene semantics into validated impedance and velocity parameters retrieved from a curated scenario database.}
    \label{fig:system_architecture}
\end{figure*}

\section{Related Work}

\paragraph{Impedance and compliance for safe interaction}
Impedance control remains the standard tool to regulate apparent stiffness and damping during contact, tracing back to Hogan’s formulation for manipulation \cite{ref2}. Subsequent work spans admittance/impedance variants, variable or learned gains, and task-specific compliance. However, most methods still rely on heuristics or local feedback rather than \emph{pre-contact} conditioning on scene semantics. In collaborative contexts, two safety families dominate: (i) \emph{power-and-force limiting} (PFL), which bounds allowable contact forces; and (ii) \emph{speed-and-separation monitoring} (SSM), which reduces speed as humans approach. ISO/TS~15066 aggregates body-region pain thresholds and guidance for allowable force/pressure under PFL, while ISO~13855 formalizes SSM via approach-speed models (e.g., $S = K T + C$) \cite{ref6,iso13855}. Our approach complements these standards by selecting impedance and speed envelopes \emph{before} execution, using task and human context instead of relying only on geometry or post-impact reactions. Empirical studies of SSM and dynamic speed control (e.g., NIST IR-7851; Byner et~al.) demonstrate the practicality of distance-based speed scaling and motivate semantics-aware scheduling layered atop these guards \cite{nist_ssm2012,byner2019ssm}.

\paragraph{Vision–language models and RAG for control}
Recent progress in VLMs has shown strong ability to interpret visual scenes and ground instructions in robot actions \cite{ref9,socratic,rt2_pmlr,openvla}. Retrieval-Augmented Generation (RAG) has further been used to constrain model outputs with curated knowledge bases, ensuring consistency and safety. In swarm robotics, ImpedanceGPT integrated VLM perception with a RAG database to schedule impedance parameters for mini-drone navigation, adapting compliance to human versus non-human obstacles in real time \cite{batool2025impedancegptvlmdrivenimpedancecontrol}. In service manipulation, Shake-VLA combined RAG with multimodal inputs for robust recipe execution despite missing or ambiguous ingredients \cite{khan2025shakevlavisionlanguageactionmodelbasedbimanual}. These examples highlight how RAG can make VLM reasoning more structured and context-aware. However, none of this work applies a VLM+RAG pipeline to human--robot interaction for humanoid manipulation. Specifically, linking semantic scene understanding to dynamic safety parameters such as impedance gains and speed has not yet been explored. Our work directly addresses this gap.

\paragraph{Egocentric semantics, proxemics, and human comfort}
Research in human-aware motion and proxemics shows that speed and proximity affect trust, comfort, and perceived safety. Higher speeds and smaller separations reduce comfort, while motion that communicates intent improves legibility and trust \cite{macarthur2017proximity,neggers2022passing,dragan2013legible,lasota2015humanaware}. Surveys on safe HRI synthesize findings across control, planning, prediction, and psychology, emphasizing the gap between geometric safeguards (SSM/PFL) and context-specific controller settings \cite{lasota2017survey,chemweno2020iso15066}. We operationalize this gap by using egocentric semantics to retrieve the least-restrictive impedance and speed envelope that remains standards-compliant for the current task and human role.

\paragraph{Positioning}
Relative to variable-impedance methods that adapt \emph{after} contact, we schedule gains and speed caps \emph{before} execution based on semantics. Relative to prior VLM applications that focus on task selection or trajectory generation, we address the missing compliance layer: semantics\,$\rightarrow$\,retrieval of safety-aligned parameters $\{\mathbf{K}_p,\mathbf{K}_d,v\}$. Geometry-based safeguards such as SSM still bound maximum speed, but semantics selects among task-appropriate envelopes (e.g., fragile handover vs.\ wipe), aiming to improve safety and comfort without collapsing to conservative global gains.

\section{System Architecture}
\subsection{Overview}
The pipeline \emph{SafeHumanoid} is organized as a modular architecture that links high-level semantic reasoning with low-level impedance control on the Unitree G1. Fig. ~\ref{fig:system_architecture} illustrates the main blocks and their interactions: (i) communication between robot and off-board server, (ii) egocentric perception Using VLM, (iii) safety parameter generation through RAG, and (iv) robot control consisting of inverse kinematics and joint-level impedance execution. This separation ensures that reasoning and control can evolve independently while remaining synchronized in real time.

\subsection{Communication Architecture}
The System modules relies on a client–server communication model between the humanoid robot and an external pc (server). The onboard computer of the robot streams egocentric images from the camera at 1–2 Hz. These data are sent to the server over TCP/IP, using Wi-Fi. The workstation processes the incoming stream and returns a validated parameter payload that includes per-joint stiffness and damping values for upper-body joints as well as a nominal velocity, in total it will return a vector of 29 values. To ensure continuity despite network delays, the onboard control loop runs at 50 Hz using the most recent valid payload, and it automatically reverts to a conservative fallback profile if communication is interrupted.
\subsection{Perception Module (VLM)}
Egocentric (RGB) frames from the robot camera are transmitted to the server, where they are processed by the Molmo-7B BnB 4-bit model. Each frame is paired with a fixed, task-specific prompt that constrains the output. The schema enforces predefined keys that cover the following: task type, main object, object fragility, human presence, obstacle count and type, workspace condition, arm posture, object location, spacing, overall complexity, and a model confidence score.

This strict prompting strategy ensures that the module directly generates semantic descriptors that are suitable for embedding and retrieval. Molmo was chosen because of its ability to interpret egocentric frames with fine-grained spatial reasoning, such as identifying whether an object is left, center, or right in the field of view. These structured outputs form the bridge between visual perception and the generation of RAG-based safety parameters described in the next section.

\subsection{Safety Parameter Generation (RAG)}
The Safety Parameter Generation module is implemented as a simple two-stage \emph{Naïve Retrieval-Augmented Generation (RAG)} system. Unlike full LLM-based RAG architectures, which use a large language model to synthesize new answers, our approach focuses on grounded retrieval from a curated scenario database, followed by deterministic formatting of the selected control parameters.  

\emph{Stage 1 — Retrieval.}  
The structured JSON output of the VLM is embedded into a 384-dimensional vector using the \texttt{all-MiniLM-L6-v2} sentence transformer. This query embedding is compared to pre-computed embeddings of all scenarios in the database using \textbf{FAISS exact nearest-neighbor search} with Euclidean distance. The database contains empirically validated scenarios, each linked to a payload of joint impedance parameters. The retrieval stage outputs the index of the top candidate together with a similarity score.  

\emph{Stage 2 — Generation.}  
Once the best-matching scenario is identified, its payload is returned in a standardized JSON/CSV structure. Each payload specifies 28 impedance parameters (14 joints $\times$ proportional and derivative gains), a nominal joint velocity, and associated metadata such as scenario type and obstacle descriptors. This guarantees consistent formatting and compatibility with the downstream control module.  

\emph{Fallback logic.}  
To ensure robustness, if two candidates have nearly identical similarity scores or if the best match falls below a predefined confidence threshold, the system rejects the retrieval and reverts to a conservative fallback parameter profile. This prevents unsafe behavior in ambiguous or novel conditions and ensures compliance with collaborative robot safety guidelines.  

In summary, the RAG module here is retrieval-driven with a lightweight generation step for structured parameter formatting, rather than a full LLM-based RAG. This design provides speed, predictability, and safety, while maintaining semantic grounding from the VLM. 

\subsection{Robot Control Module}
On-board the humanoid robot, the control module integrates inverse kinematics (IK) with a joint-space impedance controller. The IK solver operates on a reduced model where non-arm joints (legs, waist) are locked. Both left and right end-effectors are represented as operational frames, and the solver minimizes translational and rotational residuals relative to target 6-DoF poses. The optimization cost balances translation, rotation, joint regularization, and smoothness terms, with weights $\{w_{\text{trans}}, w_{\text{rot}}, w_{\text{reg}}, w_{\text{smooth}}\}$.

The solver outputs two quantities: 
\begin{enumerate}
    \item a feasible joint configuration $q_{\text{ref}}$, and
    \item feed-forward torques $\tau_{\text{ff}}$ for gravity compensation, computed using the Recursive Newton--Euler Algorithm (RNEA).
\end{enumerate}

These references are passed into the low-level impedance controller, which executes the command set
\[
\{q_{\text{ref}}, \dot q_{\text{ref}}, \tau_{\text{ff}}, K_p, K_d\}
\]
at 50~Hz on the robot internal PC. All high-frequency control remains local, ensuring stable behavior even under network delays.

For this work, predefined end-effector target poses were used as inputs to the IK solver in order to complete the pipeline demonstrations. Importantly, the architecture is source-agnostic: the same IK interface could be driven by a 6-DoF pose estimation pipeline in future work, using object pixel locations $(u,v)$ in the camera frame without changes to the downstream controller, or it could be replaced with a VLA-generated joint trajectory. Since the semantic-to-safety layer linkage is independent of the source of joint targets, these extensions are left to the future work section.

In this work, the G1’s actuators are commanded in position-control mode, which requires specifying proportional--derivative gains $(K_p, K_d)$ along with desired positions $q_{\text{ref}}$ and velocities $\dot q_{\text{ref}}$. An optional feed-forward torque $\tau_{\text{ff}}$ is also included and used for gravity compensation. The resulting motor command follows the joint-space impedance equation:
\begin{equation}
\tau = K_p \,(q_{\text{ref}} - q) + K_d \,(\dot q_{\text{ref}} - \dot q) + \tau_{\text{ff}} ,
\end{equation}
where $q$ and $\dot q$ are the measured joint positions and velocities. Because $\tau$ maps directly to end-effector interaction forces through the Jacobian,
\begin{equation}
F = J(q)^{-\top}\, \tau ,
\end{equation}
adjusting $K_p$ and $K_d$ effectively regulates the apparent stiffness and damping of the robot’s end effector. Thus, the semantic-to-safety layer of our architecture operates by scheduling $(K_p, K_d)$ profiles, which in turn modulate the interaction torques and forces experienced during contact. This establishes the direct link between high-level perception-driven reasoning and the physical compliance observed in human--robot interaction.

\section{Data Collection and Scenario Database}

To support safe and context-aware parameter retrieval, we constructed a curated dataset of manipulation scenarios performed on the Unitree G1 humanoid. The goal of this dataset is to provide empirically validated mappings between scene semantics and safe impedance settings.

\subsection{Dataset Creation}
Pilot experiments were conducted with the G1 performing tabletop manipulation tasks such as cube pickup, bottle pouring, surface wiping, and handover. Each trial was executed both in isolation and in the presence of human hands or arms to capture different safety contexts. During these sessions, the following signals were logged:
\begin{itemize}
    \item Egocentric RGB video from the RealSense camera
    \item Joint states
    \item  measured torques
    \item Proportional--derivative (PD) gains active on each joint
\end{itemize}

\subsection{Scenario Encoding}
From the collected logs, successful trials were identified based on task completion and absence of unsafe contact forces. For each validated trial, a representative snapshot was extracted and paired with the active impedance settings. These were encoded as structured entries containing:
\begin{itemize}
    \item A short natural-language description matching the VLM schema (e.g., \emph{human hand visible, cube at center, precision required})
    \item Metadata such as obstacle type and workspace condition
    \item Per-joint impedance gains $(K_p, K_d)$ for all 14 joints
    \item A nominal end-effector velocity $v$
\end{itemize}

\subsection{Database Structure}
The scenario database is stored as a single CSV file with 16 rows and 34 columns. Each row represents one validated scenario and contains the following fields:
\begin{itemize}
    \item \textbf{Identity \& semantics}
    \begin{itemize}
        \item \texttt{scenario\_id} (string), e.g., \texttt{g1\_04\_pick\_cube\_put\_in\_a\_box}
        \item \texttt{task\_enum} $\in$ \{\texttt{pick}, \texttt{handover}, \texttt{other}\}
        \item \texttt{main\_object} $\in$ \{\texttt{cube}, \texttt{fruit}, \texttt{other}\}
        \item \texttt{object\_fragility} 
        \item \texttt{human\_presence} 
        \item \texttt{nominal\_v}
    \end{itemize}
    \item \textbf{Per-joint impedance gains (28 total)} \\
    For each of the 14 joints in the left and right arms, proportional and derivative gains are stored
   
\end{itemize}

\textbf{Category coverage.} Tasks are distributed as: \texttt{pick} (9), \texttt{handover} (4), and \texttt{other} (3). Human presence conditions include \texttt{none} (9) and \texttt{hand\_visible} (7). Main objects include \texttt{cube} (3), \texttt{fruit} (3), and \texttt{other} (10).

\textbf{Parameter ranges.} Across all scenarios, proportional gains fall within $K_p \in [10, 60]$ and derivative gains within $K_d \in [0.1, 2.0]$. The database uses three discrete nominal speed $v$: slow, mid, normal speed.

\textbf{Observed profiles.} Averaged over scenarios, \texttt{fragile} objects are associated with lower stiffness (lower $K_p$), higher damping (higher $K_d$), and slower nominal velocity, whereas \texttt{non\_fragile} objects correspond to stiffer and faster profiles. These empirically tuned profiles form the payloads returned during retrieval.

\textbf{Retrieval alignment.} The fields mirror the fixed-schema JSON produced by the VLM. At runtime, the VLM output is normalized to these enums, embedded, and matched to this table’s scenarios via FAISS; the selected row’s 28 joint gains and nominal velocity form the structured parameter set returned to the robot.

\subsection{Parameter Validation}
Each candidate parameter set was tested in closed-loop execution on the G1 robot to ensure stability and compliance with safety guidelines. Validation involved two steps:
\begin{enumerate}
    \item \textbf{Stability screening.} Gains were first applied in isolated joint tests to confirm that no oscillatory or divergent behavior occurred under nominal loads.
    \item \textbf{Safety compliance.} Parameter sets were cross-checked against ISO/TS 15066 and ISO 13855 guidance. Specifically:
    \begin{itemize}
        \item For fragile objects, maximum joint stiffness was capped to maintain contact forces below published pain thresholds.
        \item For human-present scenarios, nominal speed values were restricted to those compatible with speed-and-separation monitoring standards.
        \item Scenarios that resulted in excessive force peaks or unsafe interaction behavior were excluded from the database.
    \end{itemize}
\end{enumerate}

After screening, only 16 validated scenarios were retained in the final dataset. This process ensures that every retrieval during runtime corresponds to an impedance profile that has been both empirically tested and aligned with collaborative robot safety standards.

\begin{table*}[t]
\caption{Representative outcomes of six manipulation experiments with and without human presence. Arrows indicate modulation relative to baseline.}
\label{tab:results}
\small
\setlength{\tabcolsep}{3pt}
\renewcommand{\arraystretch}{1.1}
\begin{tabularx}{\linewidth}{@{}l l l X@{}}
\toprule
\textbf{Task} & \textbf{Human} & \textbf{Adjustment} & \textbf{Outcome} \\
\midrule
Surface wipe & No & Baseline gains & Stable wiping motion \\
             & Yes & $\downarrow K_p$, $\uparrow K_d$, $\downarrow v$ & Compliant wiping near hand \\
\midrule
Pick pin (OOD) & No & Baseline gains & Successful pickup \\
               & Yes & $\downarrow K_p$, $\uparrow K_d$, $\downarrow v$ & Safe handover despite object not in dataset \\
\midrule
Pick cube (in DB) & No & Baseline gains & Successful pickup \\
                  & Yes & $\downarrow K_p$, $\uparrow K_d$, $\downarrow v$ & Safe handover, correct reversion after hand left \\
\midrule
Soy sauce bottle & Human handover & $v$: medium $\to$ slow $\to$ medium & Safe pouring with reduced motion speed \\
\bottomrule
\end{tabularx}
\end{table*}

\section{Experimental Setup}

\subsection{Hardware and Environment}
The Experiments were performed on the Unitree G1 humanoid robot equipped with an Intel RealSense RGB-D camera mounted in the head for egocentric perception and an internal PC. The onboard computer is powered by an NVIDIA Jetson Orin NX. This unit executes the inverse kinematics and low-level impedance control locally at 50 Hz.While the high-level VLM–RAG pipeline was deployed offboard on a server; a high performance PC equipped with an RTX 4090 graphics card
(24GB VRAM), communicating with the robot host via TCP/IP. This design reflects the split between heavy VLM–RAG inference (offboard) and safety-critical control (onboard). 

\begin{figure}[t] 
\centerline{\includegraphics[width=0.4\textwidth]{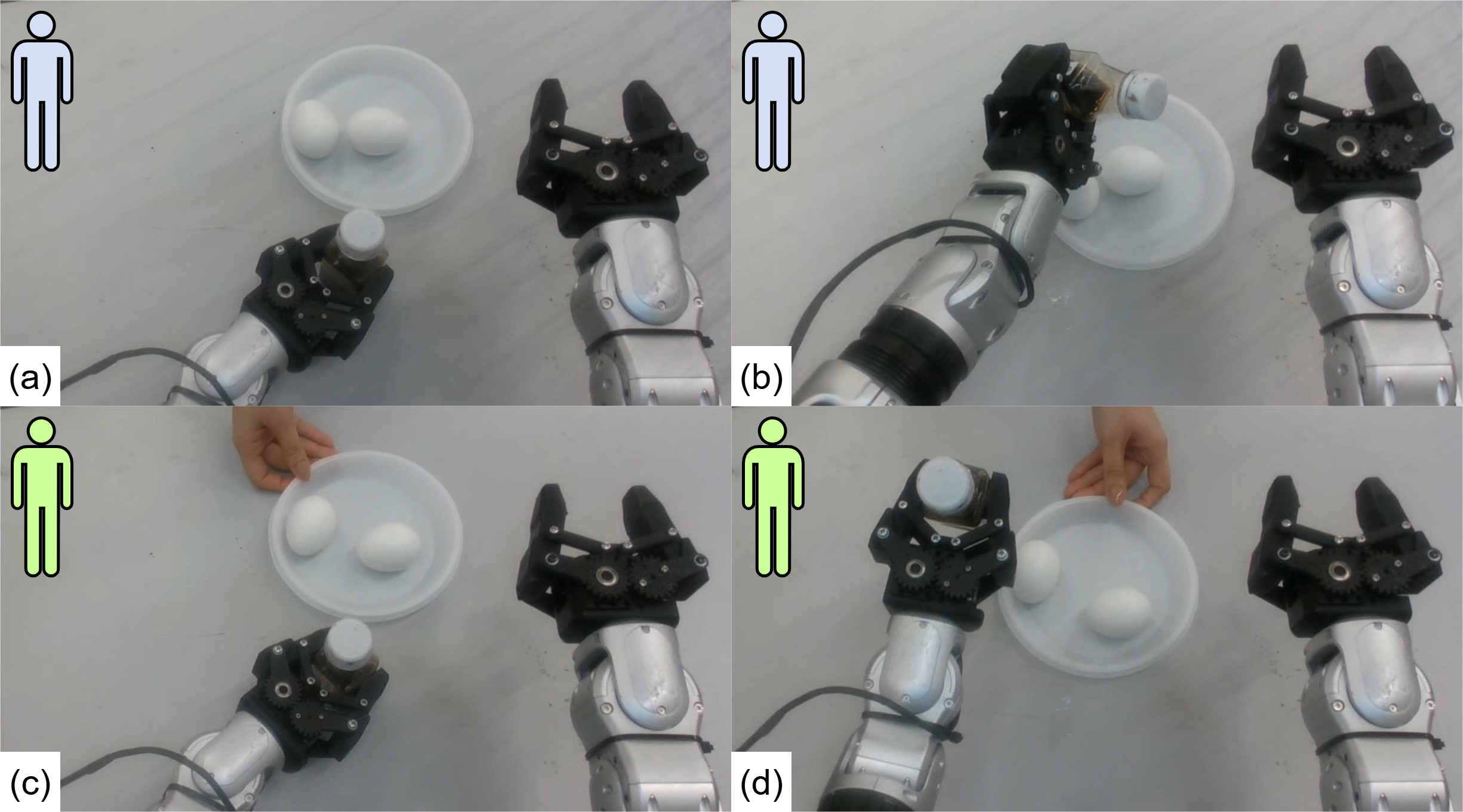}} 
\caption{Example of semantic-to-safety adaptation during a fragile object (liquid) handover. 
  (a,b) Without human presence, the system schedules moderate impedance and speed for stable handling. 
  (c,d) With human hands present, stiffness $K_p$ is reduced and damping $K_d$ is increased to ensure compliant interaction and prevent excessive contact forces.} 
\label{fig:exp} 
\vspace{-0.4cm} 
\end{figure} 

\subsection{Vision–Language Model}
Egocentric frames from the RealSense are sent to Molmo-7B on the server.Each frame is processed with a fixed JSON schema prompt that enforces structured keys: task, main object, fragility, human presence, obstacle type, workspace condition, and confidence.This ensures outputs are deterministic and compatible with the FAISS-based scenario database. Molmo was chosen for its strength in object localization and spatial reasoning.

\subsection{Procedure}
We conducted a set of manipulation experiments in a tabletop setting. Each experiment was executed in two phases: 
\begin{enumerate}
    \item \textbf{Autonomous run (no human intervention):} The robot was presented with a manipulation task (e.g., grasping a cube, pouring liquid into a cup, or handling fragile objects such as glass cup). The VLM–RAG pipeline analyzed the scene and retrieved impedance parameters $(K_p, K_d)$ and a nominal velocity $v$ aligned with the task nature and  baseline sittings.
    \item \textbf{Human-intervention run:} The same task was repeated with a human hand deliberately placed in the workspace (e.g., holding the target cup or hovering near the object). The system was expected to retrieve more compliant settings (lower $K_p$, higher $K_d$, reduced $v$) to account for human presence and safety constraints. Once the human hand left the scene, the controller was required to revert to its baseline impedance profile suitable for autonomous completion of the task.
\end{enumerate}

As shown in Fig. ~\ref{fig:exp}, we evaluated the system in a fragile-object (liquid) handover scenario. The VLM–RAG pipeline adjusted stiffness and speed depending on whether a human hand was visible, ensuring safer interaction. 
For this work, predefined 6-DoF target poses were given to the IK solver to complete the pipeline.

\subsection{Evaluation Criteria}
We defined success as proper modulation of impedance and speed: lowering stiffness near humans or fragile objects, raising stiffness where stability was needed, and returning to nominal settings once humans left. This evaluation follows the same validation logic described in the Data Collection section, aligning semantic interpretation with safe physical interaction standards.

\section{Results}
We evaluated the SafeHumanoid pipeline across six tabletop manipulation tasks. 
In all cases, the system successfully adjusted impedance parameters $(K_p, K_d)$ 
and nominal velocity $v$ according to task demands and human presence.

For surface wiping, the controller maintained baseline stiffness when no human 
was present. When a hand entered the scene, the pipeline immediately reduced $K_p$, 
increased $K_d$, and lowered velocity to produce a more compliant motion.

In the pin handover, although ``pin'' was not represented in the scenario database, 
the VLM--RAG module generalized to an appropriate compliant profile, demonstrating 
robustness to out-of-dataset objects. A similar result was observed in cube handover, 
where ``cube'' exists in the dataset: the impedance scheduling correctly switched 
to lower stiffness during human contact and returned to baseline after the human left.

For the soy sauce bottle handover, the system adjusted gains dynamically during handing over and pouring: 
nominal medium velocity was reduced to a slower profile when liquid handling was detected, 
then restored once the bottle was placed (Fig.~\ref{fig:exp}).

Across tasks, task success was maintained and impedance modulation was consistent 
with semantic context. However, latency in the offboard VLM--RAG loop reached up 
to 1.4\,s, which is unsatisfactory for dynamic HRI. This highlights the need for 
future optimization and partial onboard inference.
A summary table of representative experiments is included in Table~\ref{tab:results} provides a template of the recorded outcomes.


\section{Limitations and Future Work}
\subsection{Limitations}
While the proposed system demonstrates the feasibility of linking
semantic reasoning with impedance scheduling for safe humanoid
manipulation, several limitations remain. 

First, the offboard inference pipeline introduces non-negligible 
latency. The whole System delays of up to 1.4\,s were observed due to 
large-model inference and image transmission, restricting applicability 
to highly dynamic human--robot interaction scenarios. 

Second, the scenario database was curated manually from pilot 
experiments, requiring human supervision for task validation and 
parameter selection. This limits scalability and diversity of the database. 
Although the current set of tasks---surface wiping, pick-and-place, 
and handover---covers several fundamental human--robot interaction 
primitives, broader expansion to additional tasks and contexts is 
needed to improve generalization. 

Finally, experiments were conducted with predefined end-effector 
poses as inputs to the inverse kinematics solver. While sufficient for 
demonstrating the pipeline, this design relies on manually specified 
trajectories rather than fully autonomous action generation.
\subsection{Future Work}
Future work will address these limitations and extend the proposed framework in several directions. Reducing latency is a key priority. Possible approaches include model distillation, edge deployment of lightweight VLMs, and partial on-board inference to reduce reliance on network transmission.  

Dataset generation can be automated through simulation-to-real transfer, synthetic data augmentation, and self-labeling pipelines using depth sensing or motion capture. This will improve both the scale and coverage of the scenario database.  

Integration of depth sensing offers a promising avenue: by combining RealSense depth measurements with Molmo’s object localization, the robot could not only estimate distances between its end-effectors and nearby humans (with stricter impedance near critical regions such as the head), but also localize objects in the workspace to automate task assignment and execution. Such distance- and context-aware compliance could further increase safety and autonomy in close-contact tasks.  

Future extensions should expand semantic grounding beyond the current scene- and fragility-level context to richer physical reasoning. Instead of a simple fragile/non-fragile label, the system could estimate continuous properties such as hardness, weight, surface friction, or compliance. These attributes would allow the robot to adapt impedance more precisely — for instance, applying gentler forces to slippery or heavy objects than to stable ones. With advances in multi-modal VLMs, such property-aware safety reasoning becomes feasible, enabling more refined strategies for both human–robot and robot–object interaction.

\section{Conclusion}
In this work, we introduced \textit{SafeHumanoid}, a VLM--RAG-driven pipeline that links semantic reasoning with impedance control on a humanoid robot. Using egocentric vision, the system interprets task and human context, retrieves validated impedance and speed parameters from a curated scenario database, and integrates them with inverse kinematics and joint-level execution. 

Experiments on the Unitree G1 humanoid showed that the pipeline adapts compliance and speed in real time, producing safer and more context-aware behaviours compared to fixed-gain baselines. Notably, the approach generalized to out-of-dataset objects and tasks, while maintaining stable operation under human presence. 

Despite the observed latency of the offboard inference loop, the results highlight that semantic-to-safety grounding is a practical and extensible strategy for standard-aligned HRI. Future work will target reducing inference delays, automating dataset construction, and integrating richer multimodal reasoning to further enhance both scalability and safety.


\begin{thebibliography}{25}


\ifx \showCODEN    \undefined \def \showCODEN     #1{\unskip}     \fi
\ifx \showISBNx    \undefined \def \showISBNx     #1{\unskip}     \fi
\ifx \showISBNxiii \undefined \def \showISBNxiii  #1{\unskip}     \fi
\ifx \showISSN     \undefined \def \showISSN      #1{\unskip}     \fi
\ifx \showLCCN     \undefined \def \showLCCN      #1{\unskip}     \fi
\ifx \shownote     \undefined \def \shownote      #1{#1}          \fi
\ifx \showarticletitle \undefined \def \showarticletitle #1{#1}   \fi
\ifx \showURL      \undefined \def \showURL       {\relax}        \fi
\providecommand\bibfield[2]{#2}
\providecommand\bibinfo[2]{#2}
\providecommand\natexlab[1]{#1}
\providecommand\showeprint[2][]{arXiv:#2}

\bibitem[Ajoudani et~al\mbox{.}(2018)]%
        {ref1}
\bibfield{author}{\bibinfo{person}{Arash Ajoudani}, \bibinfo{person}{Andrea~Maria Zanchettin}, \bibinfo{person}{Serena Ivaldi}, \bibinfo{person}{Alin Albu-Sch\"{a}ffer}, \bibinfo{person}{Kazuhiro Kosuge}, {and} \bibinfo{person}{Oussama Khatib}.} \bibinfo{year}{2018}\natexlab{}.
\newblock \showarticletitle{Progress and prospects of the human---robot collaboration}.
\newblock \bibinfo{journal}{\emph{Auton. Robots}} \bibinfo{volume}{42}, \bibinfo{number}{5} (\bibinfo{date}{June} \bibinfo{year}{2018}), \bibinfo{pages}{957–975}.
\newblock
\showISSN{0929-5593}
\href{https://doi.org/10.1007/s10514-017-9677-2}{doi:\nolinkurl{10.1007/s10514-017-9677-2}}


\bibitem[Anand et~al\mbox{.}(2023)]%
        {ref3}
\bibfield{author}{\bibinfo{person}{Akhil~S. Anand}, \bibinfo{person}{Jan~Tommy Gravdahl}, {and} \bibinfo{person}{Fares~J. Abu-Dakka}.} \bibinfo{year}{2023}\natexlab{}.
\newblock \showarticletitle{Model-based variable impedance learning control for robotic manipulation}.
\newblock \bibinfo{journal}{\emph{Robotics and Autonomous Systems}}  \bibinfo{volume}{170} (\bibinfo{year}{2023}), \bibinfo{pages}{104531}.
\newblock
\showISSN{0921-8890}
\href{https://doi.org/10.1016/j.robot.2023.104531}{doi:\nolinkurl{10.1016/j.robot.2023.104531}}


\bibitem[Batool et~al\mbox{.}(2025)]%
        {batool2025impedancegptvlmdrivenimpedancecontrol}
\bibfield{author}{\bibinfo{person}{Faryal Batool}, \bibinfo{person}{Malaika Zafar}, \bibinfo{person}{Yasheerah Yaqoot}, \bibinfo{person}{Roohan~Ahmed Khan}, \bibinfo{person}{Muhammad~Haris Khan}, \bibinfo{person}{Aleksey Fedoseev}, {and} \bibinfo{person}{Dzmitry Tsetserukou}.} \bibinfo{year}{2025}\natexlab{}.
\newblock \bibinfo{title}{ImpedanceGPT: VLM-driven Impedance Control of Swarm of Mini-drones for Intelligent Navigation in Dynamic Environment}.
\newblock
\showeprint{2503.02723}


\bibitem[Brohan et~al\mbox{.}(2023)]%
        {ref7}
\bibfield{author}{\bibinfo{person}{Anthony Brohan}, \bibinfo{person}{Noah Brown}, \bibinfo{person}{Justice Carbajal}, \bibinfo{person}{Yevgen Chebotar}, \bibinfo{person}{Xi Chen}, \bibinfo{person}{Krzysztof Choromanski}, \bibinfo{person}{Tianli Ding}, \bibinfo{person}{Danny Driess}, \bibinfo{person}{Avinava Dubey}, {et~al\mbox{.}}} \bibinfo{year}{2023}\natexlab{}.
\newblock \bibinfo{title}{RT-2: Vision-Language-Action Models Transfer Web Knowledge to Robotic Control}.
\newblock
\showeprint{2307.15818}


\bibitem[Byner et~al\mbox{.}(2019)]%
        {byner2019ssm}
\bibfield{author}{\bibinfo{person}{Christoph Byner}, \bibinfo{person}{Björn Matthias}, {and} \bibinfo{person}{Hao Ding}.} \bibinfo{year}{2019}\natexlab{}.
\newblock \showarticletitle{Dynamic speed and separation monitoring for collaborative robot applications – Concepts and performance}.
\newblock \bibinfo{journal}{\emph{Robotics and Computer-Integrated Manufacturing}}  \bibinfo{volume}{58} (\bibinfo{year}{2019}), \bibinfo{pages}{239--252}.
\newblock
\showISSN{0736-5845}
\href{https://doi.org/10.1016/j.rcim.2018.11.002}{doi:\nolinkurl{10.1016/j.rcim.2018.11.002}}


\bibitem[Calinon(2016)]%
        {ref4}
\bibfield{author}{\bibinfo{person}{Sylvain Calinon}.} \bibinfo{year}{2016}\natexlab{}.
\newblock \showarticletitle{A tutorial on task-parameterized movement learning and retrieval}.
\newblock \bibinfo{journal}{\emph{Intell. Serv. Robot.}} \bibinfo{volume}{9}, \bibinfo{number}{1} (\bibinfo{date}{Jan.} \bibinfo{year}{2016}), \bibinfo{pages}{1–29}.
\newblock
\showISSN{1861-2776}
\href{https://doi.org/10.1007/s11370-015-0187-9}{doi:\nolinkurl{10.1007/s11370-015-0187-9}}


\bibitem[Chemweno et~al\mbox{.}(2020)]%
        {chemweno2020iso15066}
\bibfield{author}{\bibinfo{person}{Peter Chemweno} {et~al\mbox{.}}} \bibinfo{year}{2020}\natexlab{}.
\newblock \showarticletitle{A review of the ISO 15066 standard for collaborative robot safety}.
\newblock \bibinfo{journal}{\emph{Safety Science}}  \bibinfo{volume}{127} (\bibinfo{year}{2020}).
\newblock


\bibitem[Dragan and Srinivasa(2013)]%
        {dragan2013legible}
\bibfield{author}{\bibinfo{person}{Anca Dragan} {and} \bibinfo{person}{Siddhartha~S. Srinivasa}.} \bibinfo{year}{2013}\natexlab{}.
\newblock \showarticletitle{Generating Legible Motion}. In \bibinfo{booktitle}{\emph{RSS}}.
\newblock
\urldef\tempurl%
\url{https://personalrobotics.cs.washington.edu/publications/dragan2013legible.pdf}
\showURL{%
\tempurl}


\bibitem[Hogan(1985)]%
        {ref2}
\bibfield{author}{\bibinfo{person}{Neville Hogan}.} \bibinfo{year}{1985}\natexlab{}.
\newblock \showarticletitle{Impedance Control: An Approach to Manipulation: Part I—Theory}.
\newblock \bibinfo{journal}{\emph{Journal of Dynamic Systems, Measurement, and Control}} \bibinfo{volume}{107}, \bibinfo{number}{1} (\bibinfo{date}{march} \bibinfo{year}{1985}), \bibinfo{pages}{1--7}.
\newblock
\showISSN{0022-0434}
\href{https://doi.org/10.1115/1.3140702}{doi:\nolinkurl{10.1115/1.3140702}}


\bibitem[Huang et~al\mbox{.}(2025)]%
        {refX}
\bibfield{author}{\bibinfo{person}{Junhui Huang}, \bibinfo{person}{Yuhe Gong}, \bibinfo{person}{Changsheng Li}, \bibinfo{person}{Xingguang Duan}, {and} \bibinfo{person}{Luis Figueredo}.} \bibinfo{year}{2025}\natexlab{}.
\newblock \bibinfo{title}{ZLATTE: A Geometry-Aware, Learning-Free Framework for Language-Driven Trajectory Reshaping in Human-Robot Interaction}.
\newblock
\showeprint{2509.06031}


\bibitem[{International Organization for Standardization}(2016)]%
        {ref6}
\bibfield{author}{\bibinfo{person}{{International Organization for Standardization}}.} \bibinfo{year}{2016}\natexlab{}.
\newblock \bibinfo{booktitle}{\emph{{Robots and robotic devices -- Collaborative robots}}}.
\newblock \bibinfo{type}{Technical Specification} ISO/TS 15066:2016.
\newblock
\urldef\tempurl%
\url{https://www.iso.org/standard/62996.html}
\showURL{%
\tempurl}


\bibitem[{International Organization for Standardization}(2024)]%
        {iso13855}
\bibfield{author}{\bibinfo{person}{{International Organization for Standardization}}.} \bibinfo{year}{2024}\natexlab{}.
\newblock \bibinfo{booktitle}{\emph{Safety of Machinery—Positioning of safeguards with respect to the approach speeds of parts of the human body}}.
\newblock \bibinfo{type}{Technical Specification} ISO 13855.
\newblock
\urldef\tempurl%
\url{https://www.iso.org/standard/80590.html}
\showURL{%
\tempurl}


\bibitem[Khan et~al\mbox{.}(2025)]%
        {khan2025shakevlavisionlanguageactionmodelbasedbimanual}
\bibfield{author}{\bibinfo{person}{Muhamamd~Haris Khan}, \bibinfo{person}{Selamawit Asfaw}, \bibinfo{person}{Dmitrii Iarchuk}, \bibinfo{person}{Miguel~Altamirano Cabrera}, \bibinfo{person}{Luis Moreno}, \bibinfo{person}{Issatay Tokmurziyev}, {and} \bibinfo{person}{Dzmitry Tsetserukou}.} \bibinfo{year}{2025}\natexlab{}.
\newblock \bibinfo{title}{Shake-VLA: Vision-Language-Action Model-Based System for Bimanual Robotic Manipulations and Liquid Mixing}.
\newblock
\showeprint{2501.06919}


\bibitem[Kim et~al\mbox{.}(2024b)]%
        {ref5}
\bibfield{author}{\bibinfo{person}{Daehwa Kim}, \bibinfo{person}{Mario Srouji}, \bibinfo{person}{Chen Chen}, {and} \bibinfo{person}{Jian Zhang}.} \bibinfo{year}{2024}\natexlab{b}.
\newblock \bibinfo{title}{ARMOR: Egocentric Perception for Humanoid Robot Collision Avoidance and Motion Planning}.
\newblock
\showeprint{2412.00396}


\bibitem[Kim et~al\mbox{.}(2024a)]%
        {openvla}
\bibfield{author}{\bibinfo{person}{Moo~Jin Kim}, \bibinfo{person}{Karl Pertsch}, \bibinfo{person}{Siddharth Karamcheti}, \bibinfo{person}{Ted Xiao}, \bibinfo{person}{Ashwin Balakrishna}, \bibinfo{person}{Suraj Nair}, \bibinfo{person}{Rafael Rafailov}, \bibinfo{person}{Ethan Foster}, \bibinfo{person}{Grace Lam}, \bibinfo{person}{Pannag Sanketi}, {et~al\mbox{.}}} \bibinfo{year}{2024}\natexlab{a}.
\newblock \bibinfo{title}{OpenVLA: An Open-Source Vision-Language-Action Model}.
\newblock
\showeprint{2406.09246}


\bibitem[Lasota et~al\mbox{.}(2017)]%
        {lasota2017survey}
\bibfield{author}{\bibinfo{person}{Przemyslaw~A. Lasota}, \bibinfo{person}{Terrence Fong}, {and} \bibinfo{person}{Julie~A. Shah}.} \bibinfo{year}{2017}\natexlab{}.
\newblock \showarticletitle{A Survey of Methods for Safe Human--Robot Interaction}.
\newblock \bibinfo{journal}{\emph{Foundations and Trends in Robotics}} \bibinfo{volume}{5}, \bibinfo{number}{4} (\bibinfo{year}{2017}), \bibinfo{pages}{261--349}.
\newblock


\bibitem[Lasota and Shah(2015)]%
        {lasota2015humanaware}
\bibfield{author}{\bibinfo{person}{Przemyslaw~A. Lasota} {and} \bibinfo{person}{Julie~A. Shah}.} \bibinfo{year}{2015}\natexlab{}.
\newblock \showarticletitle{Analyzing the Effects of Human-Aware Motion Planning on Close-Proximity Human--Robot Collaboration}.
\newblock \bibinfo{journal}{\emph{Human Factors}} \bibinfo{volume}{57}, \bibinfo{number}{1} (\bibinfo{year}{2015}), \bibinfo{pages}{21--33}.
\newblock


\bibitem[Lewis et~al\mbox{.}(2020)]%
        {ref10}
\bibfield{author}{\bibinfo{person}{Patrick Lewis}, \bibinfo{person}{Ethan Perez}, \bibinfo{person}{Aleksandra Piktus}, \bibinfo{person}{Fabio Petroni}, \bibinfo{person}{Vladimir Karpukhin}, \bibinfo{person}{Naman Goyal}, \bibinfo{person}{Heinrich K\"{u}ttler}, \bibinfo{person}{Mike Lewis}, \bibinfo{person}{Wen-tau Yih}, \bibinfo{person}{Tim Rockt\"{a}schel}, \bibinfo{person}{Sebastian Riedel}, {and} \bibinfo{person}{Douwe Kiela}.} \bibinfo{year}{2020}\natexlab{}.
\newblock \showarticletitle{Retrieval-augmented generation for knowledge-intensive NLP tasks}. In \bibinfo{booktitle}{\emph{Proc. Int. Conf. on Neural Information Processing Systems}} \emph{(\bibinfo{series}{NIPS '20})}. \bibinfo{publisher}{Curran Associates Inc.}, \bibinfo{address}{Red Hook, NY, USA}, Article \bibinfo{articleno}{793}, \bibinfo{numpages}{16}~pages.
\newblock
\showISBNx{9781713829546}


\bibitem[Liang et~al\mbox{.}(2023)]%
        {ref9}
\bibfield{author}{\bibinfo{person}{Jacky Liang}, \bibinfo{person}{Wenlong Huang}, \bibinfo{person}{Fei Xia}, \bibinfo{person}{Peng Xu}, \bibinfo{person}{Karol Hausman}, \bibinfo{person}{Brian Ichter}, \bibinfo{person}{Pete Florence}, {and} \bibinfo{person}{Andy Zeng}.} \bibinfo{year}{2023}\natexlab{}.
\newblock \showarticletitle{Code as Policies: Language Model Programs for Embodied Control}. In \bibinfo{booktitle}{\emph{Proc. IEEE Int. Conf. on Robotics and Automation (ICRA)}}. \bibinfo{pages}{9493--9500}.
\newblock
\href{https://doi.org/10.1109/ICRA48891.2023.10160591}{doi:\nolinkurl{10.1109/ICRA48891.2023.10160591}}


\bibitem[Lin et~al\mbox{.}(2023)]%
        {ref8}
\bibfield{author}{\bibinfo{person}{Kevin Lin}, \bibinfo{person}{Christopher Agia}, \bibinfo{person}{Toki Migimatsu}, \bibinfo{person}{Marco Pavone}, {and} \bibinfo{person}{Jeannette Bohg}.} \bibinfo{year}{2023}\natexlab{}.
\newblock \showarticletitle{Text2Motion: from natural language instructions to feasible plans}.
\newblock \bibinfo{journal}{\emph{Auton. Robots}} \bibinfo{volume}{47}, \bibinfo{number}{8} (\bibinfo{date}{Nov.} \bibinfo{year}{2023}), \bibinfo{pages}{1345–1365}.
\newblock
\showISSN{0929-5593}
\href{https://doi.org/10.1007/s10514-023-10131-7}{doi:\nolinkurl{10.1007/s10514-023-10131-7}}


\bibitem[MacArthur et~al\mbox{.}(2017)]%
        {macarthur2017proximity}
\bibfield{author}{\bibinfo{person}{Keith~R. MacArthur}, \bibinfo{person}{Kimberly Stowers}, {and} \bibinfo{person}{P.~A. Hancock}.} \bibinfo{year}{2017}\natexlab{}.
\newblock \showarticletitle{Human-Robot Interaction: Proximity and Speed---Slowly Back Away from the Robot!}. In \bibinfo{booktitle}{\emph{Advances in Human Factors in Robots and Unmanned Systems}}, \bibfield{editor}{\bibinfo{person}{Pamela Savage-Knepshield} {and} \bibinfo{person}{Jessie Chen}} (Eds.). \bibinfo{address}{Cham}, \bibinfo{pages}{365--374}.
\newblock


\bibitem[Neggers et~al\mbox{.}(2022)]%
        {neggers2022passing}
\bibfield{author}{\bibinfo{person}{M.M.E. Neggers}, \bibinfo{person}{others Margot M. E.~Neggers}, \bibinfo{person}{Raymond~H. Cuijpers}, \bibinfo{person}{Peter A.~M. Ruijten}, {and} \bibinfo{person}{Wijnand~A. IJsselsteijn}.} \bibinfo{year}{2022}\natexlab{}.
\newblock \showarticletitle{The effect of robot speed on comfortable passing distances}.
\newblock \bibinfo{journal}{\emph{Frontiers in Robotics and AI}}  \bibinfo{volume}{9} (\bibinfo{year}{2022}).
\newblock


\bibitem[Szabo et~al\mbox{.}(2012)]%
        {nist_ssm2012}
\bibfield{author}{\bibinfo{person}{Sandor Szabo}, \bibinfo{person}{William Shackleford}, \bibinfo{person}{Richard Norcross}, {and} \bibinfo{person}{Jeremy Marvel}.} \bibinfo{year}{2012}\natexlab{}.
\newblock \bibinfo{booktitle}{\emph{A Testbed for Evaluation of Speed and Separation Monitoring in Shared Workspaces}}.
\newblock \bibinfo{type}{{T}echnical {R}eport} NIST IR 7851. \bibinfo{institution}{NIST}.
\newblock
\href{https://doi.org/10.6028/NIST.IR.7851}{doi:\nolinkurl{10.6028/NIST.IR.7851}}


\bibitem[Zeng et~al\mbox{.}(2022)]%
        {socratic}
\bibfield{author}{\bibinfo{person}{Andy Zeng} {et~al\mbox{.}}} \bibinfo{year}{2022}\natexlab{}.
\newblock \showarticletitle{Socratic Models: Composing Zero-Shot Multimodal Reasoning via Language}. In \bibinfo{booktitle}{\emph{ICLR}}.
\newblock
\urldef\tempurl%
\url{https://socraticmodels.github.io/}
\showURL{%
\tempurl}


\bibitem[Zitkovich et~al\mbox{.}(2023)]%
        {rt2_pmlr}
\bibfield{author}{\bibinfo{person}{Brianna Zitkovich}, \bibinfo{person}{Tianhe Yu}, \bibinfo{person}{Sichun Xu}, \bibinfo{person}{Peng Xu}, \bibinfo{person}{Ted Xiao}, \bibinfo{person}{Fei Xia}, \bibinfo{person}{Jialin Wu}, \bibinfo{person}{Paul Wohlhart}, \bibinfo{person}{Stefan Welker}, \bibinfo{person}{Ayzaan Wahid}, {et~al\mbox{.}}} \bibinfo{year}{2023}\natexlab{}.
\newblock \showarticletitle{{RT-2}: Vision-Language-Action Models Transfer Web Knowledge to Robotic Control}. In \bibinfo{booktitle}{\emph{Proc. of The 7th Conference on Robot Learning}}.
\newblock
\urldef\tempurl%
\url{https://proceedings.mlr.press/v229/zitkovich23a/zitkovich23a.pdf}
\showURL{%
\tempurl}


\end{thebibliography}
\end{document}